\documentclass[conference]{IEEEtran}
\IEEEoverridecommandlockouts

\usepackage{cite}
\usepackage{amsmath,amssymb,amsfonts}
\usepackage{algorithmic}
\usepackage{graphicx}
\usepackage{textcomp}
\usepackage{xcolor}
\usepackage{rotating}
\usepackage{tabularray}
\usepackage{subcaption}
\usepackage{siunitx}
\DeclareSIUnit{\kWh}{kWh}  
\DeclareSIUnit{\Wh}{Wh}  
\usepackage{csquotes}
\usepackage{tablefootnote}
\usepackage{booktabs}
\usepackage[nolist]{acronym}
\usepackage{pifont}
\usepackage{hyperref}

\usepackage{xcolor}
\usepackage{amsmath,amssymb}

\definecolor{janecolor}{rgb}{0.2,0.6,0.6}
\definecolor{johncolor}{rgb}{0,0.7,0}

\definecolor{todocolor}{rgb}{0.9,0.1,0.1}
\definecolor{changedcolor}{rgb}{0.42,0.27,0.57}
\definecolor{addedcolor}{rgb}{0.867,0.176,0.361}

\newcommand{\redacted}[1]{\emph{[anonymized for review]}}

\newcommand{\xmark}{\ding{55}}%
\newcommand{\ra}[1]{\renewcommand{\arraystretch}{#1}}
\def\BibTeX{{\rm B\kern-.05em{\sc i\kern-.025em b}\kern-.08em
    T\kern-.1667em\lower.7ex\hbox{E}\kern-.125emX}}
\begin{document}

%

\title{Guidelines for the Quality Assessment \linebreak{}of Energy-Aware NAS Benchmarks}
\author{\IEEEauthorblockN{Nick Kocher\IEEEauthorrefmark{1}\IEEEauthorrefmark{2}, Christian Wassermann\IEEEauthorrefmark{1}, Leona Hennig\IEEEauthorrefmark{3}, Jonas Seng\IEEEauthorrefmark{4},\\
Holger Hoos\IEEEauthorrefmark{1}\IEEEauthorrefmark{5}, Kristian Kersting\IEEEauthorrefmark{4}, Marius Lindauer\IEEEauthorrefmark{3} and Matthias Müller\IEEEauthorrefmark{1}}
\IEEEauthorblockA{\IEEEauthorrefmark{1}RWTH Aachen University}
\IEEEauthorblockA{\IEEEauthorrefmark{3}Leibniz University Hannover}
\IEEEauthorblockA{\IEEEauthorrefmark{4}Technical University Darmstadt}
\IEEEauthorblockA{\IEEEauthorrefmark{5}Leiden University}
\IEEEauthorblockA{\IEEEauthorrefmark{2}Corresponding Author: kocher@aim.rwth-aachen.de}}

\maketitle

\begin{abstract}
\ac{NAS} accelerates progress in deep learning through systematic refinement 
of model architectures. The downside is increasingly large energy consumption during the search process. 
Surrogate-based benchmarking mitigates the cost of full training by querying a pre-trained surrogate to obtain an estimate for the quality of the model. Specifically, energy-aware benchmarking aims to make it possible for \ac{NAS} to favourably trade off model energy consumption against accuracy. 
Towards this end, we propose three design principles for such energy-aware benchmarks:
(i) reliable power measurements, 
(ii) a wide range of GPU usage, and 
(iii) holistic cost reporting. 
We analyse EA-HAS-Bench based on these principles and find that the choice of GPU measurement API has a large impact on the quality of results. 
Using the Nvidia \ac{SMI} on top of its underlying library influences the sampling rate during the initial data collection, returning faulty low-power estimations. 
This results in poor correlation with 
accurate measurements obtained from an external power meter.
With this study, we bring to attention several key considerations when performing energy-aware surrogate-based benchmarking and derive first guidelines that can help design novel benchmarks.
We show a narrow usage range of the four GPUs attached to our device, ranging from \qty{146}{W} to \qty{305}{W} in a single-GPU setting, and narrowing down even further when using all four GPUs.
To improve holistic energy reporting, we propose calibration experiments over assumptions made in popular tools, such as Code Carbon, thus achieving reductions in the maximum inaccuracy from \qty{10.3}{\%} to \qty{8.9}{\%} without and to \qty{6.6}{\%} with prior estimation of the expected load on the device.

\end{abstract}


\section{Introduction}
\label{sec:intro}
Deep learning has accelerated progress in many machine-solvable tasks such as image recognition \cite{Lec98LeNet}, image segmentation \cite{Lon15FCN} and natural language processing \cite{Hoc97LSTM}. The trend away from feature engineering towards architecture engineering has given rise to 
increasingly large neural network models.
Starting from the breakthrough of AlexNet \cite{Kriz12AlexNet} at the ImageNet competition in 2012, various different architectures, including GoogLeNet \cite{Sze15GoogleNet}, ResNet \cite{16HeResNet} and the Transformer \cite{Vas17Transformer}, have introduced 
a plethora of ways to bundle and stack different layers of neural networks. 
In contrast to these manual engineering approaches, neural architecture search (\ac{NAS}), 
a research area within automated machine learning (AutoML) \cite{Hut2019AutoML}, aims to automate the complex process
of finding
best-performing architecture for a given task
\cite{Els19NASsurvey}. \ac{NAS} has contributed to frontier models in image classification \cite{PHA21EfficientNet} and natural language processing \cite{So21Primer}.

While architectures get more refined, their energy demand can quickly skyrocket during the long searches on large numbers of GPUs \cite{Zop17NasCost}. The increase in search cost, as well as overall high training and inference cost, sparked research efforts towards Green AI and Green AutoML \cite{SCH20GAI,Tor23GAutoML}. 
Beside environmental considerations, Green AI additionally advocates for more efficient AI research, 
in order to democratise research for smaller-scale and publicly funded institutions. 

One way to mitigate high computational requirements for participation in research
is through the use of 
more efficient
benchmarking techniques. 
Surrogate-based benchmarks, such as NAS-Bench-201 \cite{Dong2020NAS-Bench-201}, 
replace expensive function evaluations (\emph{i.e.}, fully training a candidate neural network architecture) by proxy evaluations at almost negligible cost.
Here, instead of a full training cycle, we only need to query the (pre-trained) surrogate model to retrieve (an estimate of) 
the performance associated with the given candidate architecture.

Even though only querying the surrogate helps decrease the overall cost of benchmarking new NAS approaches, we are ultimately interested in finding architectures that are themselves energy-efficient during training and inference.
Hardware-aware \ac{NAS} (HW-NAS) promises to identify
architectures at the Pareto front of energy efficiency and accuracy \cite{Ben21HWNas}. 
This has important applications in the context of
energy-constrained devices, such as battery-powered cars or low-power devices in the Internet of Things (IoT) \cite{DBLP:journals/pieee/ZhouCLZLZ19}, \cite{DBLP:journals/network/LiOD18}.

\begin{figure*}[ht]
    \centering
    \includegraphics[width=2\columnwidth]{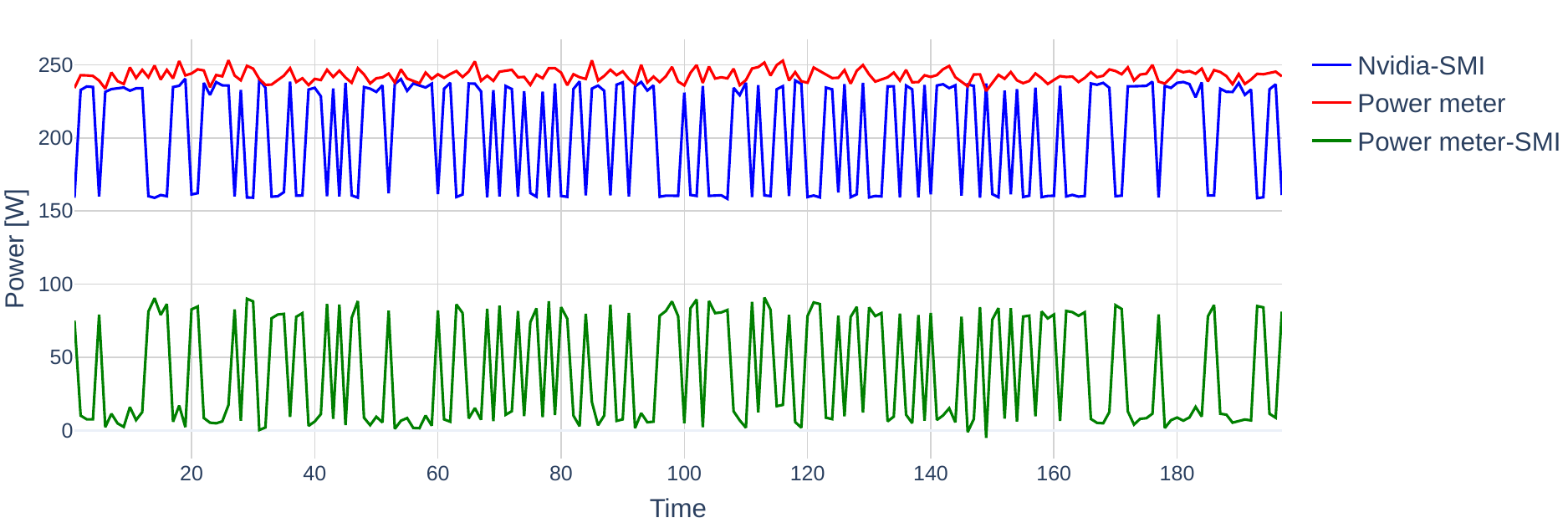}
    \caption{Switching of low- and high-power states measured by \ac{SMI} queries during training of a neural architecture on one GPU. The red line is the power measured by the external power meter. The blue line is the power measured by \ac{SMI} and the green line is the difference between the two.}
    \label{fig:intro image}
\end{figure*}

To facilitate the multi-objective search for architectures,
HW-NAS benchmarks provide additional surrogates for specific hardware \cite{Li21HWNasbench} and may resort to reusing performance metrics \cite{Dong2020NAS-Bench-201}.
The Energy-aware Hyperparameter and Architecture Search Benchmark (EA-HAS-Bench) aims to construct
a hardware-agnostic search space based on energy measured using Nvidia \ac{SMI} \cite{Dou23EAHAS}. As shown
in \autoref{fig:intro image}, this may lead to inaccurate
measurements.
For these hardware-agnostic energy-aware \ac{NAS} benchmarks to work in complementarity with other hardware-aware \ac{NAS} benchmarks, we define three key design principles:

    \textbf{Conduct reliable power measurements.}
    Most modern \ac{NAS} methods employ some form of multi-fidelity optimisation based on partial information (\emph{i.e.,} based on training for a small number of epochs or on a subset of data). For these methods to work, we need energy measurements that are already reliable at low fidelity.

    \textbf{Allow for a wide range of GPU usage.}
    The search space or the proposed datasets need to be diverse enough to include high- and low-power ML models on a single GPU.
    The benchmark should produce meaningful results for multiple different use cases. A search that only uses 100W on a 800W Nvidia H100 wastes expensive hardware resources, while architectures resulting from a 800W-search will be meaningless
    for smaller IoT devices.

    \textbf{Report holistic model cost.}
    Energy costs of using the final model should include the cost for the complete training device, 
    \emph{i.e.}, not only processors, accelerators and memory,
    during training and inference. This is especially important for battery-constrained IoT devices, where reliable cost information is crucial for the battery life estimation of the device.

\subsection*{Main contributions}
Our main contribution is a large-scale study of the EA-HAS-Bench data collection scheme, which we introduce in \autoref{sec:prelimstudy}.
Based on the previously introduced design principles and using different power measurement strategies, we show that: 
\begin{enumerate}
    \item Nvidia \ac{SMI} produces \textbf{poor correlation} to external power meter measurements on a per-epoch basis. This often results
    in a mix of high- and low-power states as shown
    in \autoref{fig:intro image}.
    \item Measurements using \ac{SMI} produce \textbf{insufficient samples in the low-power epochs} to correctly estimate the energy consumption during those epochs. 
    In contrast
    to their published code, in their original work, Dou et al.
    used pyNVML to measure the GPU energy consumption \cite{25NVINVML}. With the pyNVML-based setup,
    we were able to reproduce the data collection without error.
    \item EA-HAS-Bench has a low GPU usage
    range, but high correlation with the power meter at full fidelity. 
    \item Holistic energy measurements tools such as Code Carbon \cite{Cou24CodeCarbon} underestimate the energy consumption during training, and we propose a method for an offline calibration of the non-measured consumption.
\end{enumerate}

\section{Related work}




With the remarkable success of deep learning techniques across a wide range of applications,
the problem of finding optimal architectures for the task at hand has rapidly 
become important to solve. Pioneering approaches in \ac{NAS} include the utilisation of Bayesian optimisation \cite{Zim21AutoPytorch}, reinforcement learning (RL) \cite{Zop17NasCost, pham2018ENAS} and evolutionary algorithms (EA) \cite{Elsken2018EfficientMN, real2017EvolutionaryNAS}, as well as the introduction of differentiable neural architectures that enable
the optimisation of architectures using stochastic gradient descent (SGD) \cite{liu2018darts, Chu2020DARTSRS, Chu2019FairDARTS, Chen2019ProgressiveDARTS, Zela2019UnderstandingDARTS}. While RL and EA-based \ac{NAS} methods often cover a more flexible macro design space, they come with
high resource consumption. In contrast, gradient-based approaches focus on micro-level architecture decisions but are significantly more efficient;
this is a consequence of treating the search space as continuous instead of discrete, thereby training only one super-network with continuous decisions on components. 
Thus, the problem of finding an optimal architecture can be decomposed into two distinct levels, where in the inner optimisation, network weights are trained with current architecture components, and in the outer optimisation, the architecture components are updated based on  network performance. Overall efficacy is then further enhanced by approximations of the first-order solution of this bi-level optimisation problem.
To improve the efficiency of \ac{NAS} algorithms, recent approaches make use of training-free evaluation scores to guide the search through the space of architectures~\cite{Li2023ZeroShotNAS, Qiao2024MicroZeroShotNAS, Zhang2023AnEM}.

To foster the development, reproducibility, and fair comparison of \ac{NAS} algorithms, a number of different benchmarks have been proposed. 
These NAS benchmarks define a search space over the architectures and train all (or a large subset of) candidate architectures to  yield performance metrics 
such as validation and test accuracy. 
Popular examples of such benchmarks include NAS-Bench-101/201/301~\cite{Ying19nasbench101, Dong2020NAS-Bench-201, Siems2020NASBench301AT} and JAHS Bench~\cite{Ban22jahsbench201}. Since solving \ac{NAS} problems comes with significant consumption of compute resources (and thus electric power), energy-aware \ac{NAS} benchmarks have been proposed 
to obtain similar benefits as from existing \ac{NAS} benchmarks in the more challenging setting of developing energy-efficient multi-fidelity \ac{NAS} methods. Popular benchmarks include the Energy-aware Hyperparameter and Architecture Search Benchmark (EA-HAS-Bench)~\cite{Dou23EAHAS} and the Energy Consumption-aware NAS benchmark (EC-NAS)~\cite{Bak24ECNASBench}. EA-HAS-Bench and EC-NAS 
aim to be hardware-agnostic. However, there are benchmarks that do consider hardware properties to foster the development of NAS algorithms that search for architectures tailored to certain hardware configurations~\cite{Li21HWNasbench, fu2021autoNBA}.

\section{Power measurement tools}
In this section, we describe the different power measurement tools we used for our experiments. 

\subsection{External power meter}
To validate the node-internal power measurements, a calibrated ZES ZIMMER LMG450\footnote{\url{https://www.zes.com/en/Products/Precision-Power-Analyzers/LMG450}} power meter was connected to the four \acp{PSU} of the compute node.
Experiments were performed on the same node running Rocky Linux 9.3 using two Intel(R) Xeon(R) Platinum 8\,480 processors with 112 cores in total, 2\,000 GB of memory, and four H100 GPUs attached to the node.
The LMG450 has a measurement accuracy of $\qty{0.07}{\percent} + \qty{0.04}{\percent}$ of the measuring range.
We record the continuously integrated active power of all four channels as measured by the internal shunt current sensors.
The sampling frequency was configured to \qty{100}{\milli\second}, while the current and voltage ranges were set to \qty{16}{\ampere} and \qty{250}{\volt}, respectively.

\subsection{Nvidia \ac{SMI}}
The Nvidia \ac{SMI} is a command-line utility providing management and monitoring functionality for GPU statistics. 
Power is measured through the 
\texttt{$<$power.draw$>$}
call, 
which 
by default outputs the average power draw over \qty{1}{\second} estimated by the GPU itself
\cite{24YanSMIAttention,25NVISMI}. This can lead to significantly under-reported 
consumption, especially during short kernel executions \cite{24YanSMIAttention}. 
Steady power loads have a margin of error of $5\%$ on the H100 \cite{24YanSMIAttention}.
The \ac{NVML} is the library underlying  \ac{SMI} \cite{25NVINVML}. It also provides a Python binding through \texttt{pip} (called pyNVML) and is used for the tools such as Code Carbon; For more details, see \autoref{sec:codecarbon}.

\subsection{RAPL}

Intel's \ac{RAPL} technology manages the power consumption of the underlying processor \cite{How10RAPL}. As such, it estimates the energy consumption of certain so-called power domains. 
On most modern Intel processors, starting from the Sandybridge architecture, (at least) the \texttt{PKG} and \texttt{DRAM} domains are available, which manage the entire CPU socket and the memory, respectively. For ease of understanding, we will call them CPU and memory consumption going forward.

These domains are accessible through \acp{MSR}, typically exposed in the power cap interface of the Linux kernel. We use
the PyRAPL library \cite{Rou23pyRAPL} to import the estimations into the benchmark directly through the provided Python API. Other tools, such as PAPI \cite{McC14PAPI} and LIKWID \cite{Trei10Likwid}, are popular alternatives to obtain high-level access to the power cap interface. 
\ac{RAPL} is generally considered to be accurate on server architectures developed after Sandybridge; the main reason for this is the switch from  power modeling \cite{How10RAPL} using performance counters to a measurement-based approach \cite{Des16RAPLAcc}.
Note that to reproduce results from our study, 
elevated access or ownership of the power cap interface may be required.

\subsection{Code Carbon}\label{sec:codecarbon}
Reliable tools for quantifying and reporting emissions from machine learning models are essential for both research and environmental impact assessments \cite{DBLP:journals/corr/abs-1910-09700}, \cite{DBLP:journals/corr/abs-1911-08354}. Code Carbon is such an energy reporting tool specifically designed to track carbon emissions from computational processes by monitoring energy use and regional energy mix in $g\mathit{CO}_{2}\mathit{eq}$, grams of $\mathit{CO}_2$ equivalents \cite{Cou24CodeCarbon}. Grams of $\mathit{CO}_2eq$ quantify the carbon footprint of ML models by accounting for their energy consumption and grid carbon intensity, using the global warming potential (GWP) to standardise emissions from different greenhouse gases.
For electricity consumption during computing, this measure is simply based on the proportional amount of $CO_2$ emitted on the power grid, given an appropriate regional energy mix, in our case $43\%$ fossil fuel, resulting in a carbon intensity of $\qty{381}{g CO_{2}/k Wh}$. \footnote{\url{https://github.com/bundesAPI/smard-api}}

We use Code Carbon as a proxy for
for measuring energy consumption, as it is a widely adopted tool; yet, its reliability requires further assessment \cite{DBLP:journals/corr/abs-2306-08323}.
Its main routine is split into two parts. In the first stage, the energy consumption of the executed code block is modelled as
\begin{equation} \label{eq:code carbon energy}
    E_{code carbon} = (E_{NVML} + E_{CPU} + \hat{E}_{memory}) \cdot \mathrm{PUE},
\end{equation}
where memory consumption $\hat{E}_{memory}$ is estimated using \qty{3}{\watt} per \qty{8}{\giga\byte} of memory. $E_{NVML}$ and $E_{CPU}$ are the energy consumptions reported by \ac{NVML} and for the CPU domain in \ac{RAPL}, respectively.
The node energy consumption is then scaled by the \ac{PUE}.
The \ac{PUE} represents a factor incorporating the additional energy used to operate a compute cluster; an industry average \ac{PUE} of $1.58$ has been reported for 2020 \cite{Pat22Footprint}.
In the second stage, energy consumption is transformed to $\mathrm{CO}_2eq$ 
emissions by multiplying the average carbon intensity ($\mathrm{CO}_2eq$ per \unit{\kWh}) of the local power grid with the energy consumption.
In our study, we manually extract the energy consumption prior to scaling with the \ac{PUE}, since it is constant across all experiments and not used by \ac{SMI} or the external power meter.

\section{Fair measurements} \label{sec:measurement}
The main objective of using a power meter in this context is to empirically evaluate
how to perform measurements at scale and to improve existing measurement tools. 
The main complication of using the power meter in our setup is that it measures the energy consumption of the entire node and not merely  that attributable to the model training on the GPU.
To be able to obtain a fair comparison between \ac{SMI} and the power meter, the latter needs to be calibrated to a base power consumption; everything above this base level would then be counted towards actual consumption by the benchmark. 
In \autoref{eq:naive} below, we present a na\"{\i}ve way to think about capturable power consumption $P_{naive}$ on a compute node. We can obtain $P_{idle}$ from $P_{naive}$ by measuring an idle run using the power meter and subtracting \ac{RAPL} and \ac{SMI} measurements.
During the actual experiments, this information could then be used to obtain the GPU power consumption as measured by the power meter.
\begin{align}
    P_{naive} &= P_{CPU} + P_{Memory} + P_{GPU} + P_{idle} \label{eq:naive}
\end{align}

For the idle run, $P_{idle}$ was calculated as \qty{783}{\watt}. 
However, when running a stress test such as Firestarter \cite{Hac13Firestarter} to maximise load on the CPU and memory, we obtained $P_{idle} = $ \qty{941}{\watt}. This leads us to conclude that there is an uncaptured energy-using component on a node during load; this is made explicit in \autoref{eq:busy1} below. We thus need to determine a $P_{busy}$ that approximates the uncaptured power consumption from the CPU and memory to obtain unbiased GPU consumption during the experiments; see \autoref{eq:busy2}.
\begin{align}
    P_{total} &=P_{naive} + P_{uncaptured} 
    \label{eq:busy1} \\
    &= P_{CPU} + P_{Memory} + P_{GPU} + P_{busy} \label{eq:busy2}
\end{align}

Thankfully, initial experiments show that calculating large prime numbers with a $P_{busy}$ of \qty{811}{\watt} is a good lower bound for the busy power consumption. 

Our procedure for calculating the GPU power consumption as measured by the power meter thus looks as follows:
\begin{equation}
P_{GPU} = P_{total} - P_{CPU} - P_{Memory} - P_{busy}
\label{eq:gpu_only}
\end{equation}
When comparing the power meter to the measurements from \ac{SMI}, we use this adjusted power consumption.

\section{Validation Study} \label{sec:prelimstudy}
The EA-HAS-Bench(mark) provides surrogate models for a complex RegNet search space. RegNets are a variation of ResNets, where residual blocks are replaced by recurrent neural networks, such as LSTMs \cite{Jin23RegNet}. 
The data for the surrogates is gathered by randomly sampling both traditional hyperparameters, as well as architectural details, such as the number of residual blocks from a large search space, and then training the resulting model while measuring the energy consumption using \ac{SMI}. As claimed by Dou et al.,
these surrogates show a Pearson correlation coefficient of 0.89 with the collected data. Validating the surrogates is beyond the scope of our study; 
instead, we are interested in the collected raw energy data.
Specifically, we performed a large-scale study of 500 sampled architectures to understand the accuracy of the above described data collection scheme underlying the benchmark.

\subsection{Setup of validation experiments}
We mimic the procedure by randomly sampling architectures from the RegNet \cite{Jin23RegNet} search space and then sequentially train the resulting models on Tiny Imagenet.
Tiny Imagenet is a smaller version of the original Imagenet dataset \cite{Rus15ImageNet}
that contains 1\,000\,000 images of 200 classes with 500 images per class. Images are downsized to $64 \times 64 \times 3$. This is the largest dataset used in the EA-HAS-Benchmark.
We modified the training code of the benchmark to obtain the energy consumption of the entire node from the power meter in addition to the already contained \ac{SMI} samples. The benchmark samples \ac{SMI} with a rate of \qty{10}{\hertz} (one sample each \qty{100}{\milli\second}) from a separate thread. While this is configurable, we used the default of \qty{10}{\hertz}, as the power meter was also storing samples at this frequency.  

The power meter output was recorded on a
separate external machine, and results were stored after experiments had been concluded. The measurements started and stopped before and after each training run. Therefore,  additionally measuring energy with the power meter did not induce any load on the CPU while the architectures were being trained. To obtain the adjusted energy consumption in postprocessing, we also sampled the \ac{RAPL} values at the end of each epoch using PyRAPL, as outlined in \autoref{sec:measurement}.
For this validation study, we only used one out of four available GPUs.

\subsection{Validation results}
\begin{figure}[t]
    \centering
    \includegraphics[width=\columnwidth]{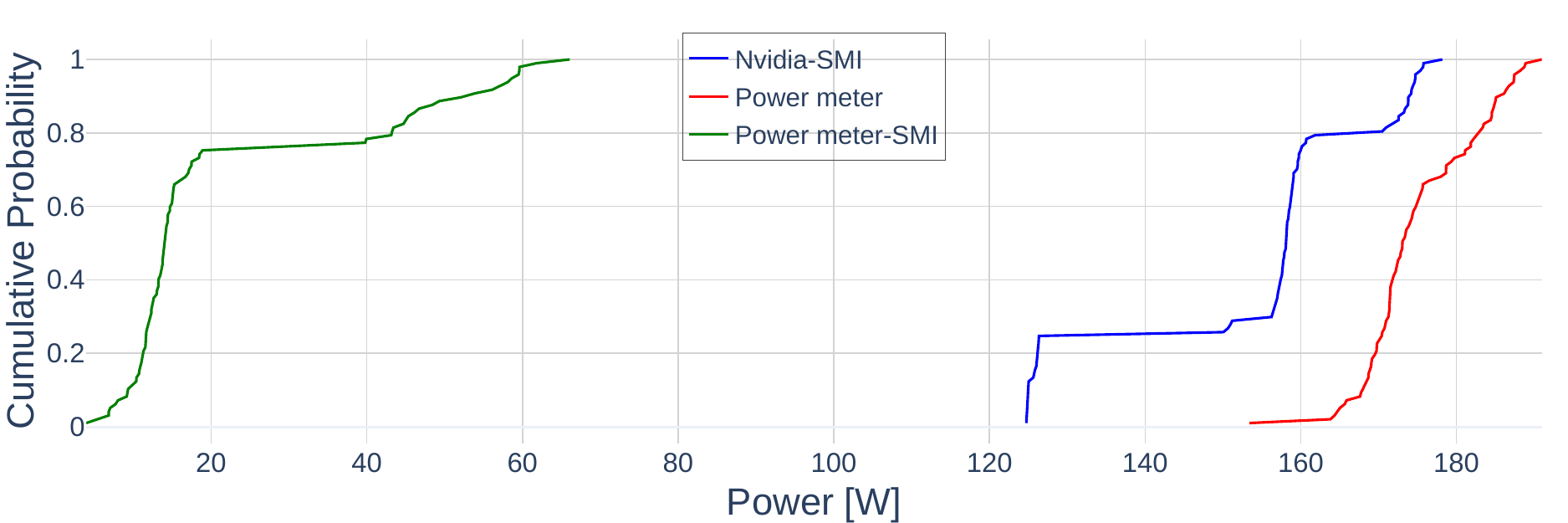}
    \caption{\ac{eCDF} of the power meter measurements and the \ac{SMI} sampled power estimates for an example model training on one GPU.}
    \label{fig:sample_bias_ecdf}
\end{figure}

Epochs measured in this way exhibit a poor energy correlation with the power meter, with a Pearson correlation coefficient of 0.64. When aggregating epochs for the full training of a sampled model, the coefficient improves to 0.99.
The comparison between full training and per-epoch measurements is highlighted in the first row of \autoref{table:results}.
Throughout the training of the models in this study, the per-epoch measurements are split between low- and high-power states on the GPU, with a plateau of non-measured power states in the \ac{eCDF}; see \autoref{fig:sample_bias_ecdf} as an example. In this specific case, the model had no measured power consumption between \qty{130}{\watt} and \qty{150}{\watt}. 
While the \ac{eCDF} of the power consumption per epoch appears to be strictly monotonically increasing
across all training runs,
we observed such a split behaviour for almost all of the individual training runs. 
The reason for this behaviour is the low sampling rate in $39\%$ of the sampled epochs. In these epochs, there are no more than 10 power samples measured during an average training time of \qty{11.8}{\second}. 

\begin{figure}[t]
    \centering
    \includegraphics[width=\columnwidth]{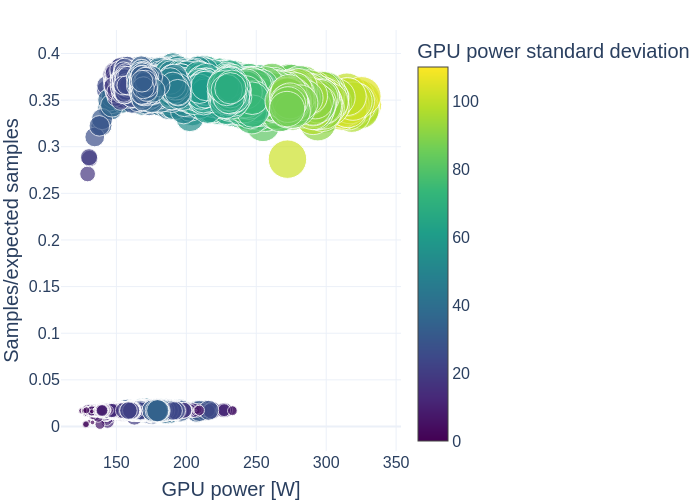}
    \caption{GPU power vs the ratio of taken samples and expected samples. We see that there are no high power states with a low sample ratio.}
    \label{fig:gpu_samples}
\end{figure}

The divide between low and high sampling rates is visualised in \autoref{fig:gpu_samples}. In fact, we did not observe any high GPU power epochs with a low number of taken samples in the total number of 53\,850 measured epochs. When leaving these epochs out of consideration, the Pearson correlation coefficient increased to 0.95; see the second row of \autoref{table:results}. Overall, the Pearson correlation coefficient between the number of samples per epoch and reported energy usage per epoch was calculated as 0.57. Correcting for low number of samples turned the correlation around to -0.39. Ideally, these values should be uncorrelated. 
Explaining the reasons for this undesired behaviour is beyond the scope of our study, although we observed in subsequent experiments that it is due to the use of \ac{SMI} on top of pyNVML.

\section{Main study}\label{sec:mainstudy}

In this section, we first discuss the setup for three different experiments we conducted as a follow-up to the initial study, and then present and discuss results from these three experiments, 
particularly focusing on the three design principles introduced in \autoref{sec:intro}.

\subsection{Setup of experiments}\label{sec:setup}

All three experiments used the same general protocol as the initial study and were executed on the same hardware. 
The follow-up experiments, in which we sampled 20 architectures per experiment, enrich the data collection by using additional power measurement tools and training procedures. 
In all of the experiments, we measured the energy consumption using Code Carbon in addition to the power meter to understand the quality of the energy cost reporting. The custom measuring interval of \qty{100}{\milli\second} in Code Carbon deviates from the default of \qty{15}{\second} to match the power meter and the benchmark sampling speed. In the second small-scale experiment, all four H100s were used and the model was trained in a distributed fashion. In the third experiment, we replaced SMI with pyNVML to analyse the low sampling rates reported in the previous section. pyNVML was used in a previous version of the EA-HAS-Bench to perform the data collection, and was, to the best of our knowledge, the method used in the original paper of Dou et al. \cite{Dou23EAHAS}. We performed the third experiment twice, once only with pyNVML enabled, to confirm the cause of the behaviour from \ac{SMI} showcased in the previous section, and a second time also with the power meter and Code Carbon enabled.

\subsection{Reliable power measurements}\label{sec:results}
We now discuss the results from measuring energy consumption using different tools during the training of RegNets as part of the data collection phase of an energy-aware benchmark. In \autoref{table:results}, we present the main aggregated statistics from all experiments. The table includes the Spearman and Pearson correlation coefficients between the energy measured by the tested tool and the power meter, as well as the result of Kolmogorov-Smirnov (KS) test with 
$\alpha = 0.05$. Our null hypothesis was that the measurements obtained via a given tool follow the same distribution as the 
ground truth from the power meter. 
\begin{table}[t] 
\centering 
\ra{1.3}
\resizebox{\columnwidth}{!}{ 
\begin{tabular}{@{}lrrrcrrrcrrr@{}}\toprule 
& \multicolumn{3}{c}{Epoch} & \phantom{abc}& \multicolumn{3}{c}{Full Training}  \\
\cmidrule{2-4} \cmidrule{6-8} 
& Spearman\textsuperscript{\dag} & Pearson\textsuperscript{\dag} & KS\textsuperscript{*} && Spearman & Pearson & KS\\ \midrule
\multicolumn{1}{l}{Single GPU}  \\ 
\hspace{0.5cm}SMI & 0.58 & 0.64 & \xmark && 0.99 & 0.99 & \xmark \\
\hspace{0.5cm}SMI (corrected) & 0.99 & 0.95 & \checkmark && 0.99 & 0.99 & \xmark \\
\hspace{0.5cm}NVML & 0.98 & 0.99 & \checkmark && 0.99 & 0.99 & \checkmark \\
\hspace{0.5cm}Code Carbon & 0.99& 0.99& \xmark&& 0.99& 0.99& \checkmark\\ \cline{1-1}
\multicolumn{1}{l}{Multi GPU}\\
\hspace{0.5cm}SMI & 0.88& 0.97& \xmark&& 0.99& 0.99& \checkmark\\
\hspace{0.5cm}Code Carbon & 0.93& 0.99& \xmark&& 0.99& 0.99& \checkmark\\
\bottomrule
\multicolumn{8}{l}{\textsuperscript{*} KS: Kolmogorov–Smirnov test, $\alpha = 0.05$, \textsuperscript{\dag} correlation coefficients} \\%
\end{tabular}}
\caption{Correlation of measurements obtained via different tools with the power meter measurements.}
\label{table:results}
\end{table}

We tested \ac{SMI} and Code Carbon for both single-GPU and multi-GPU training, and we present aggregated results for both single epochs and the full training cycle. This includes the results from the preliminary study. Additionally, the results for the single-GPU measurements from \ac{SMI} corrected for sufficiently sampled epochs and \ac{NVML} are depicted. For the Code Carbon measurements, we observe an overall high correlation with the power meter measurements for single epochs, both during the single- and multi-GPU run. This is very similar to the results obtained via \ac{NVML}. With Code Carbon using the pyNVML API, this further provides evidence towards SMI being the cause for the previously described low sample rate.
The measurement quality for SMI improved significantly compared to the preliminary results, when sampling during a multi-GPU run. The Pearson correlation coefficient jumps to 0.97 for single epochs. 
Only the NVML and corrected SMI distributions pass the Kolmogorov-Smirnov test and are thus statistically indifferent from those 
obtained from the power meter. The measurements on full training runs all correlate well with the power meter measurements. 

The Kolmogorov-Smirnov tests for full training show that all measurements but \enquote{\ac{SMI} on a single GPU} are not significantly differently distributed from the power meter measurement. We note that there are many more full training cycles for the single-GPU-SMI setting, 
since the large-scale study was performed exclusively using this setting.
\begin{figure}[t]
    \centering
    \includegraphics[width=\columnwidth]{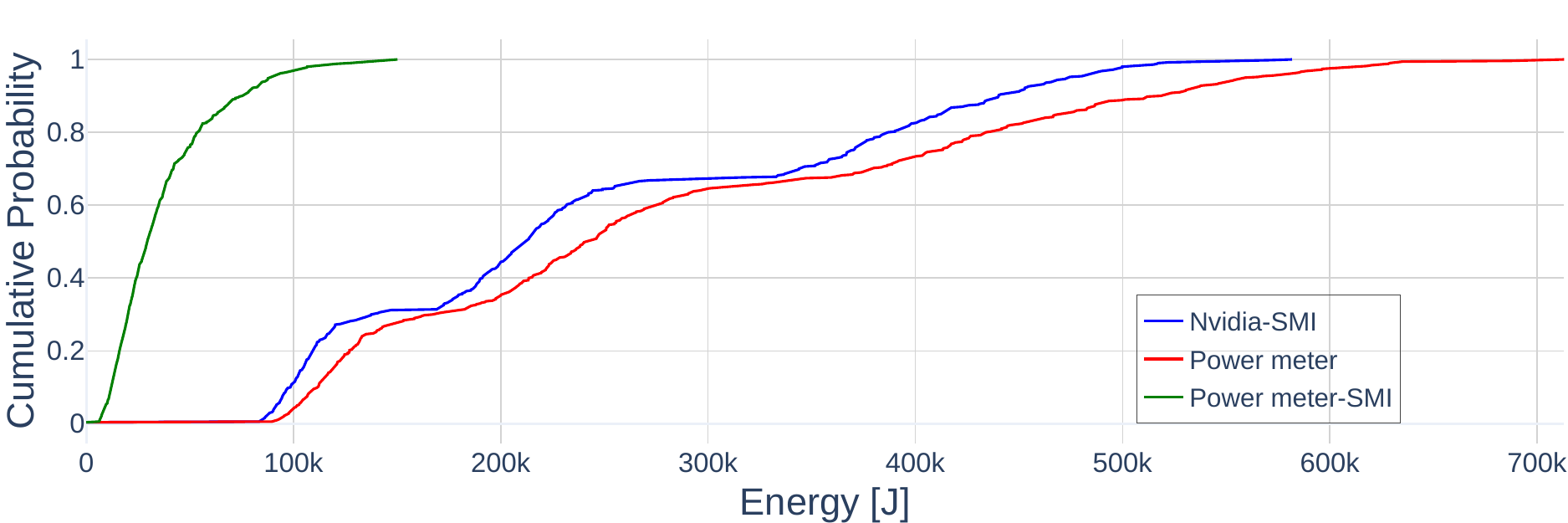}
    \includegraphics[width=\columnwidth]{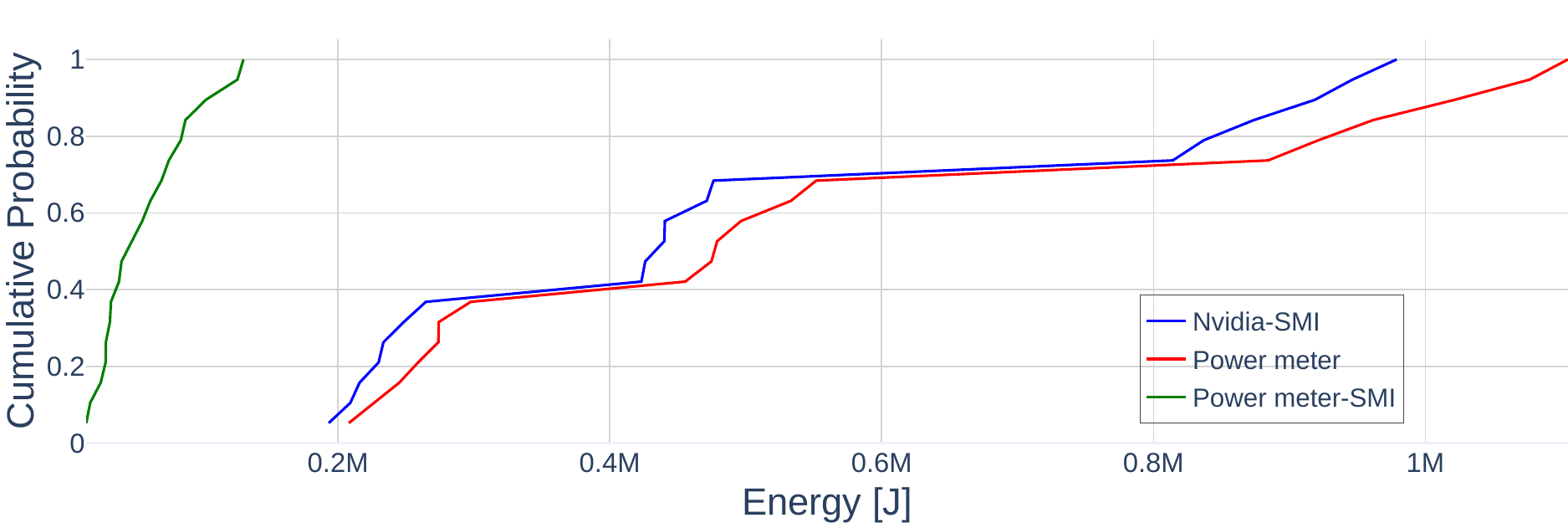}
    \caption{Top: \ac{eCDF} of the aggregated single-GPU training for \ac{SMI} and power meter measurements.
    Bottom: \ac{eCDF} of the aggregated multi-GPU training for \ac{SMI} and power meter measurements.}
    \label{fig:energy_model_SMI}
\end{figure}
The similarity between the \ac{SMI} measurements on single- and on multi-GPU is also highlighted in \autoref{fig:energy_model_SMI}. Here, we visualise the \ac{eCDF}s over the full training energy costs for the single- and multi-GPU scenarios, respectively. According to both distribution functions, \ac{SMI} reports consistently less energy than the power meter. The median training cost doubles from \qtyrange{67}{132}{\Wh}, while the median inaccuracy from \ac{SMI} increases from \qtyrange{8.3}{12.2}{\Wh}.
\begin{figure*}[h!]  
    \centering
    \begin{subfigure}[t]{0.45\textwidth}
        \centering
        \includegraphics[width=\textwidth]{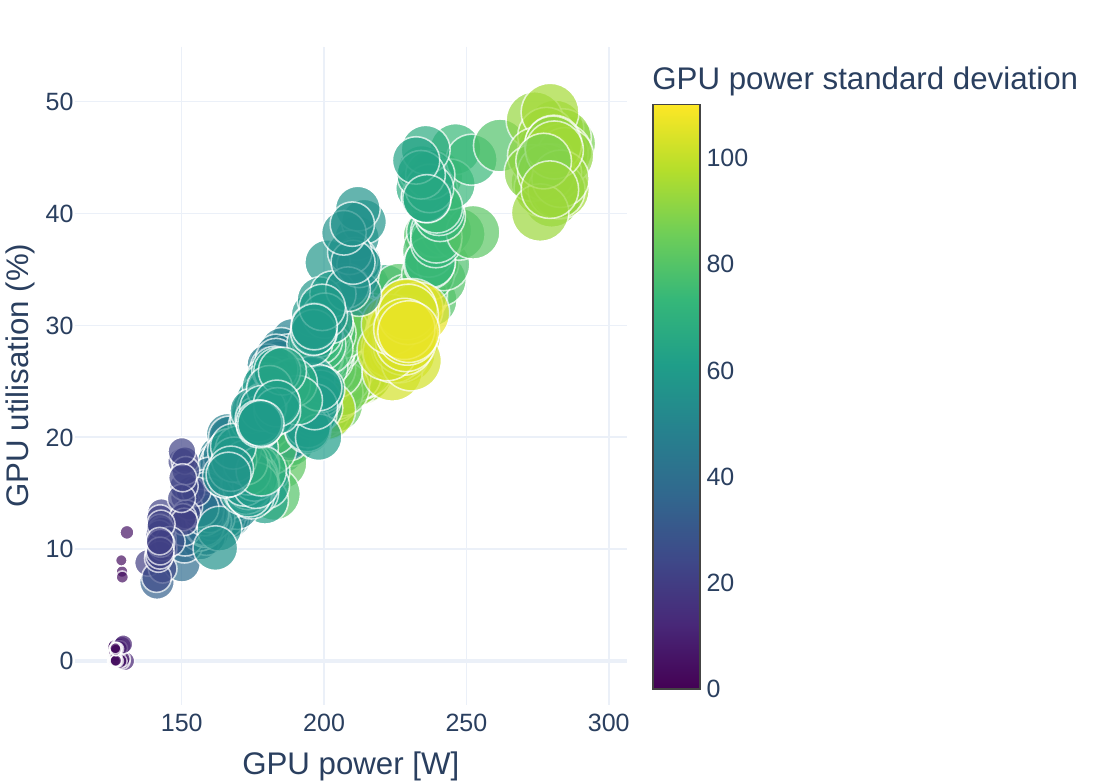}
        \includegraphics[width=\textwidth]{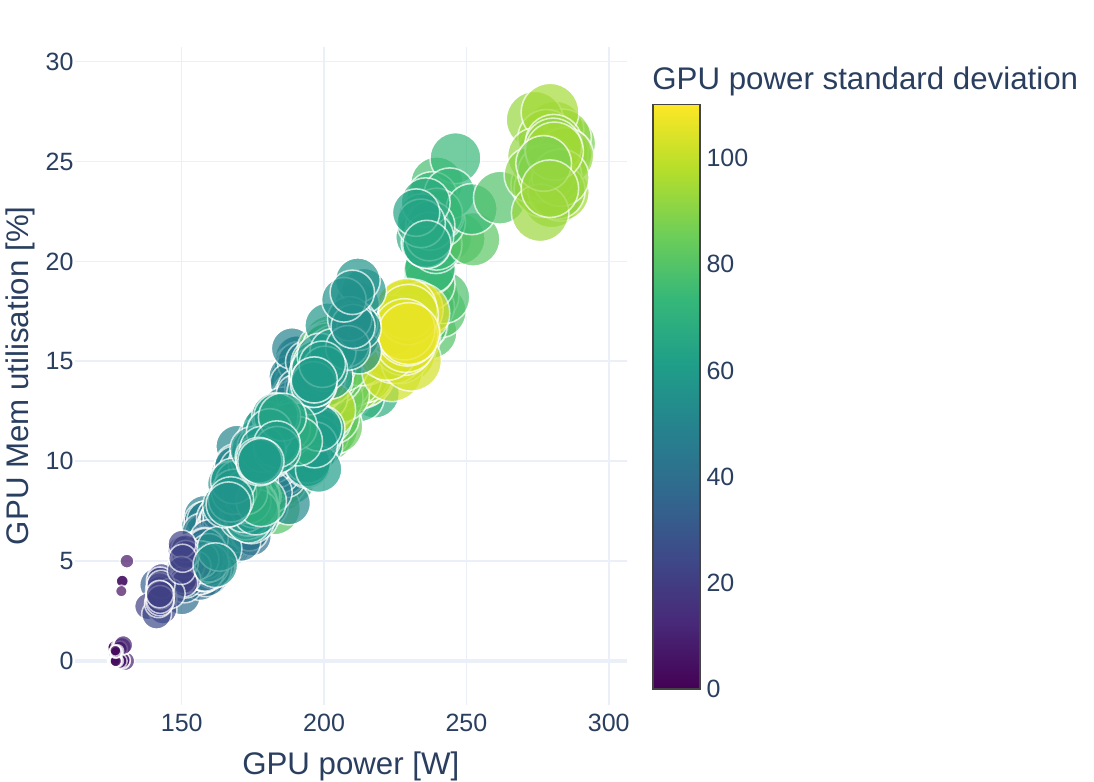}
        
    \end{subfigure}
    \hfill
    \begin{subfigure}[t]{0.45\textwidth}
        \centering
        \includegraphics[width=\textwidth]{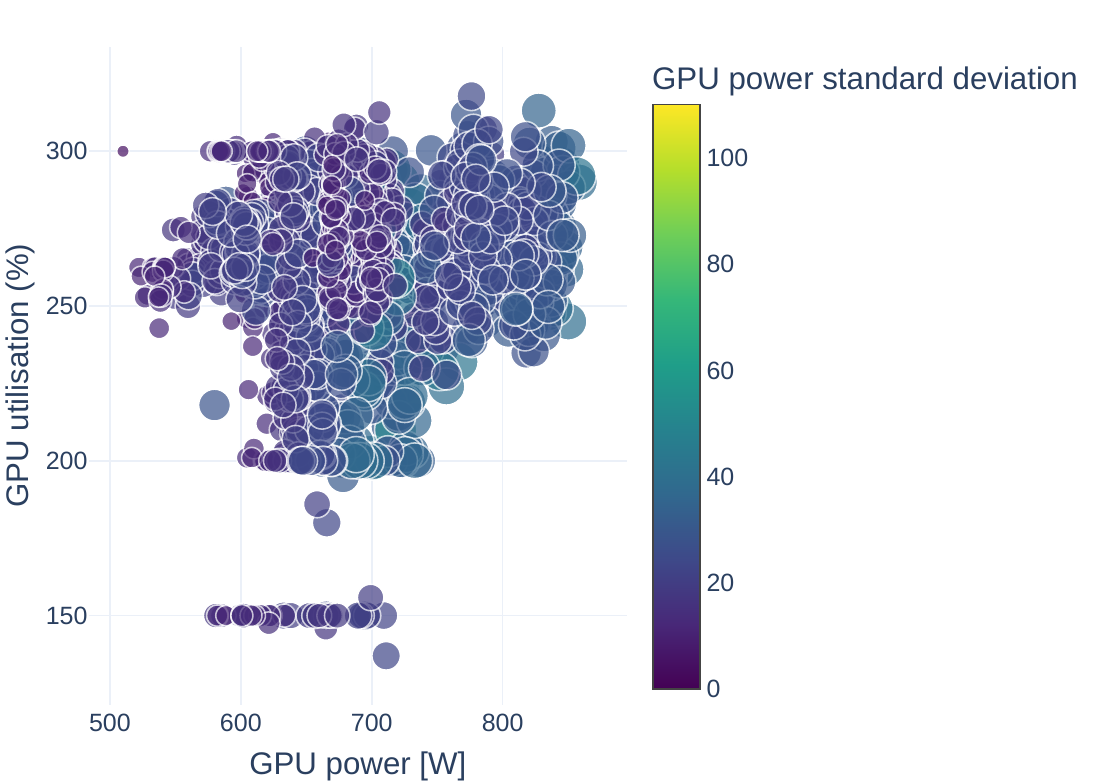}
        \includegraphics[width=\textwidth]{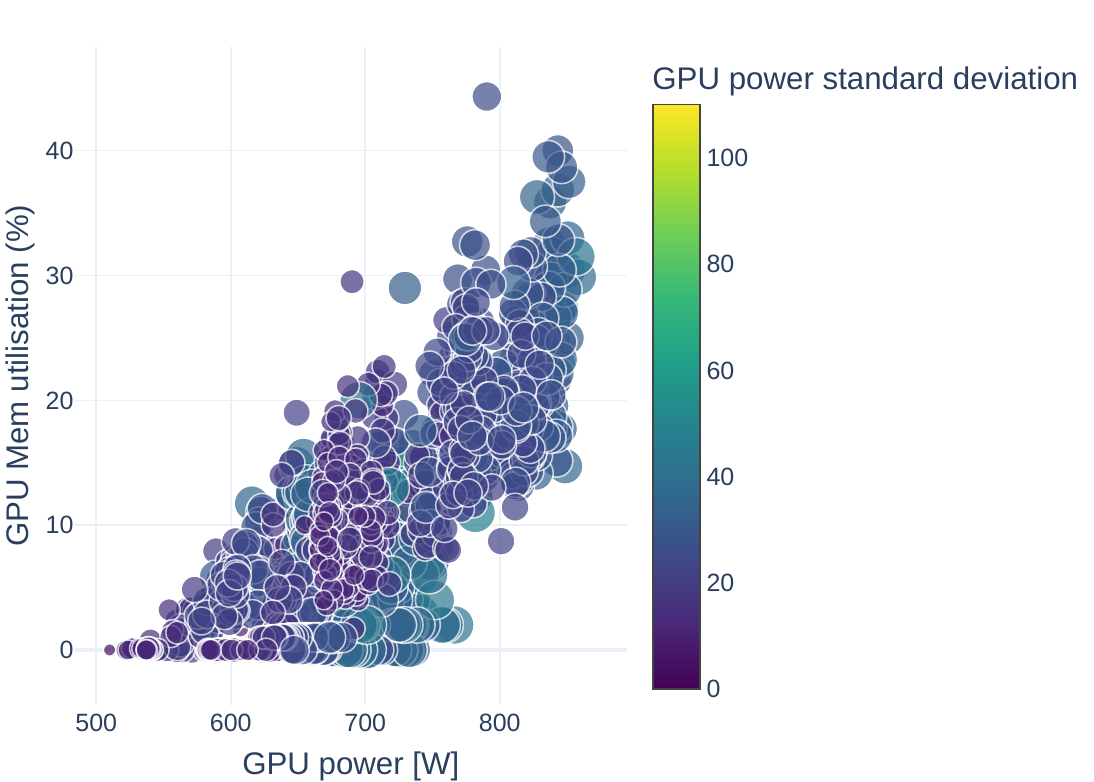}

    \end{subfigure}
    \hfill
    \caption{Top left: GPU power consumption vs \emph{GPU utilisation} for \emph{single-GPU} training. Bottom left: GPU power consumption vs \emph{GPU memory utilisation} for \emph{single-GPU} training. Top right: GPU power consumption vs \emph{GPU utilisation} for \emph{multi-GPU} training. Bottom right: GPU power consumption vs \emph{GPU memory utilisation} for \emph{multi-GPU} training. All data is aggregated across epochs. Brighter colours indicate higher standard deviation.}
    \label{fig:gpu_correlation}
\end{figure*}
\subsection{Range of GPU usage}

The range of different power draws on the GPU is relatively narrow, with \qtyrange{146}{305}{\watt}  compared to the base consumption of the GPU of \qty{75}{\watt} and the maximum consumption of \qty{800}{\watt}.
The GPU was never fully used, with maximum usage around $40\%$. This is a sign that the RegNet search space is not ideal for the H100 GPU. Testing on additional hardware is out of scope of this study, but would be worthwhile investigating in future work.
\autoref{fig:gpu_correlation} shows the linear correlation between the energy and usage reported by \ac{SMI} for all epochs measured during the single-GPU measurements. Brighter colours and larger circles indicate higher standard deviations during measurements. We observe that the higher variance also increases almost linearly with the measured power consumption. 
The multi-GPU measurements on the righthand side of \autoref{fig:gpu_correlation} demonstrate different behaviour. The power draw of all four GPUs is uncorrelated to the GPU usage, with a Pearson correlation coefficient of -0.03. The power consumption in this setting is
only correlated with the memory usage of the GPUs.

\subsection{Reporting of holistic energy costs}
\begin{figure}[h!]
    \centering
    \includegraphics[width=\columnwidth]{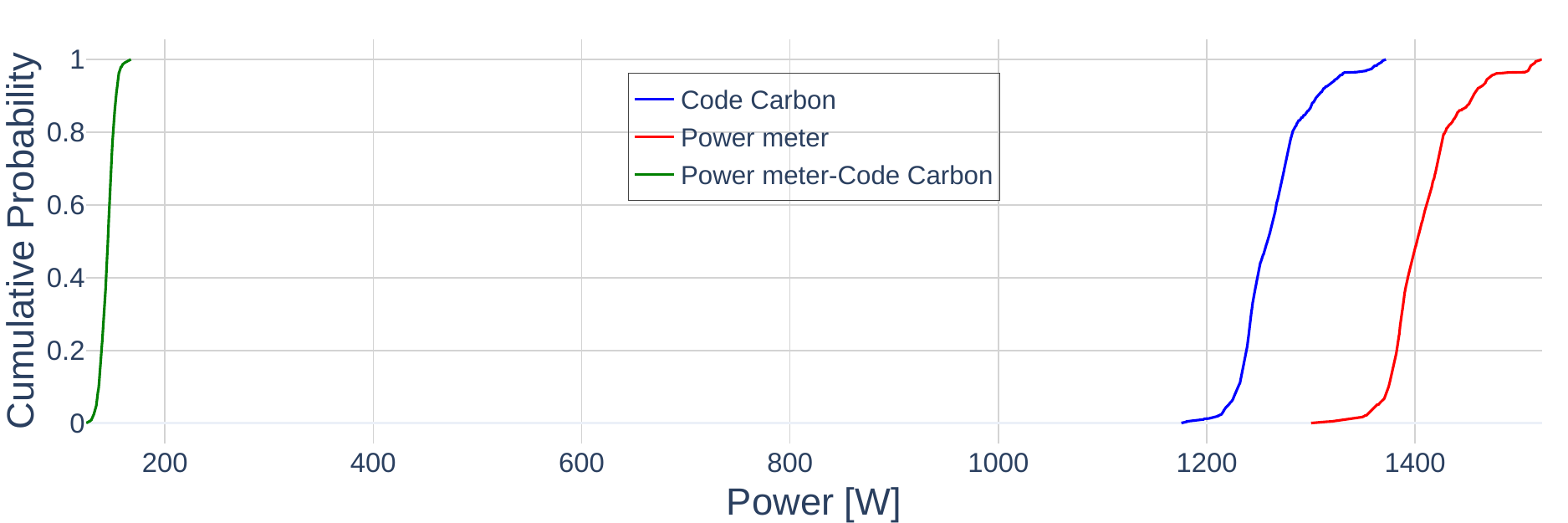}
    \caption{\ac{eCDF} plot for the epoch power measurements aggregated across all training runs for Code Carbon and the power meter.}
    \label{fig:epoch_power_code_carbon}
\end{figure}

Dou et al. \cite{Dou23EAHAS} claim that EA-HAS-Bench is "$\textit{aware of the overall search energy cost}$". The statistical indifference between the distributions of (corrected) \ac{SMI}/\ac{NVML} measurements and the power meter was shown in \autoref{table:results}. We briefly bring attention back to the calculations we had to perform to extract only the GPU power from the power meter, as described in \autoref{eq:gpu_only} in \autoref{sec:measurement}. While GPU measurements are accurate, the associated consumption does not comprise the majority of the energy cost during training, even during multi-GPU training.
Therefore, we analysed whether the accurate measurements could be supplemented by publicly available cost reporting tools, such as Code Carbon, to make the benchmark aware of the \textit{overall} energy cost.
\begin{figure}[h!]
    \centering
    \includegraphics[width=\columnwidth]{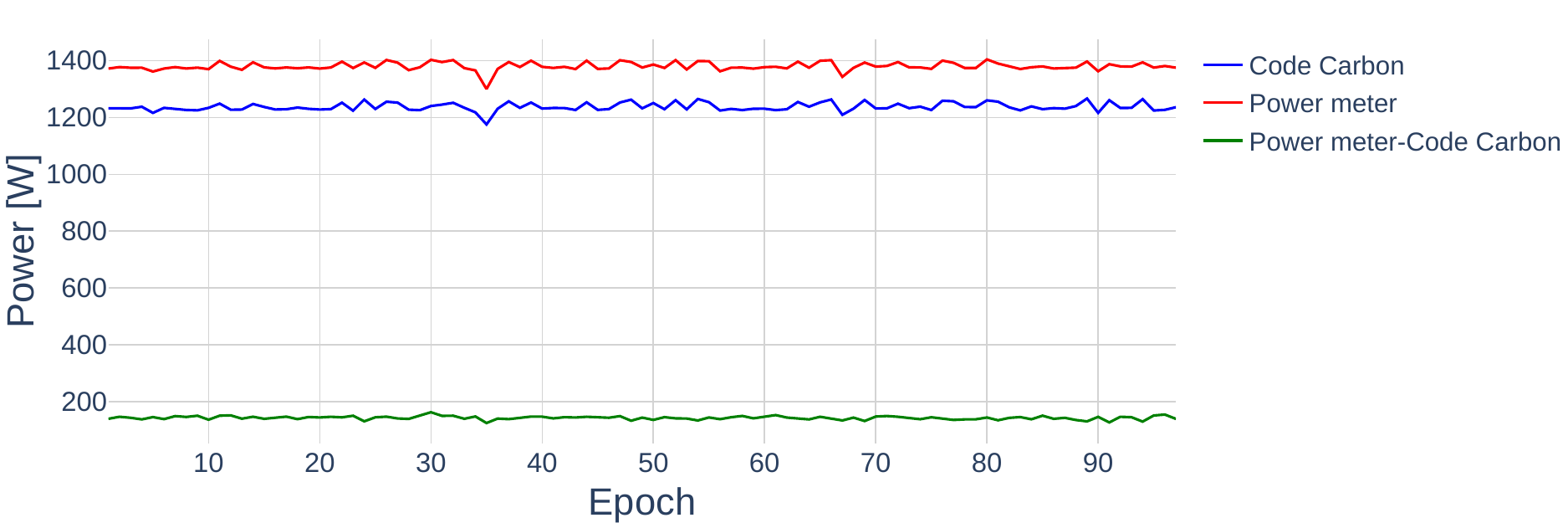}
    \caption{Power measurements across time for Code Carbon and the power meter for an example model training on one GPU.}
    \label{fig:code_carbon_example}
\end{figure}
In \autoref{sec:codecarbon}, we described the calculation procedure for Code Carbon. To summarise, Code Carbon uses \ac{NVML}, \ac{RAPL} and an estimate of the memory power consumption, in our case \qty{750}{\watt}. With the real memory power consumption of our node being \qty{12}{\watt}, the overestimation of memory consumption by RAPL helps compensate for the lack of off-socket power consumption on the node. In \autoref{table:results} we see that the measurements from Code Carbon  are linearly correlated with those obtained from the power meter. In \autoref{fig:code_carbon_example}, we visualise for one example model training the constant power difference between Code Carbon and the power meter. This is again highlighted in \autoref{fig:epoch_power_code_carbon}, this time aggregated across the entirety of the second experiment. Compared to the SMI data, we observe a relatively constant underestimation of the power consumption by Code Carbon.

Instead of assuming large amounts of available memory, we base the calculations on empirically evaluated data. To this end, we propose to use a sensible range of estimations between the idle power consumption and the busy power consumption from Firestarter. The range of estimations should be provided by system administrators through the baseboard management controller available on most modern servers. These systems have relatively low sampling rates, usually once every 5 minutes; however, 
a constant load does not require high sampling rates. If more precise bounds are needed, calibrating with a custom load, such as our prime number calculation, would be required. 

\begin{figure*}[t]
    \centering
    \includegraphics[width=\textwidth]{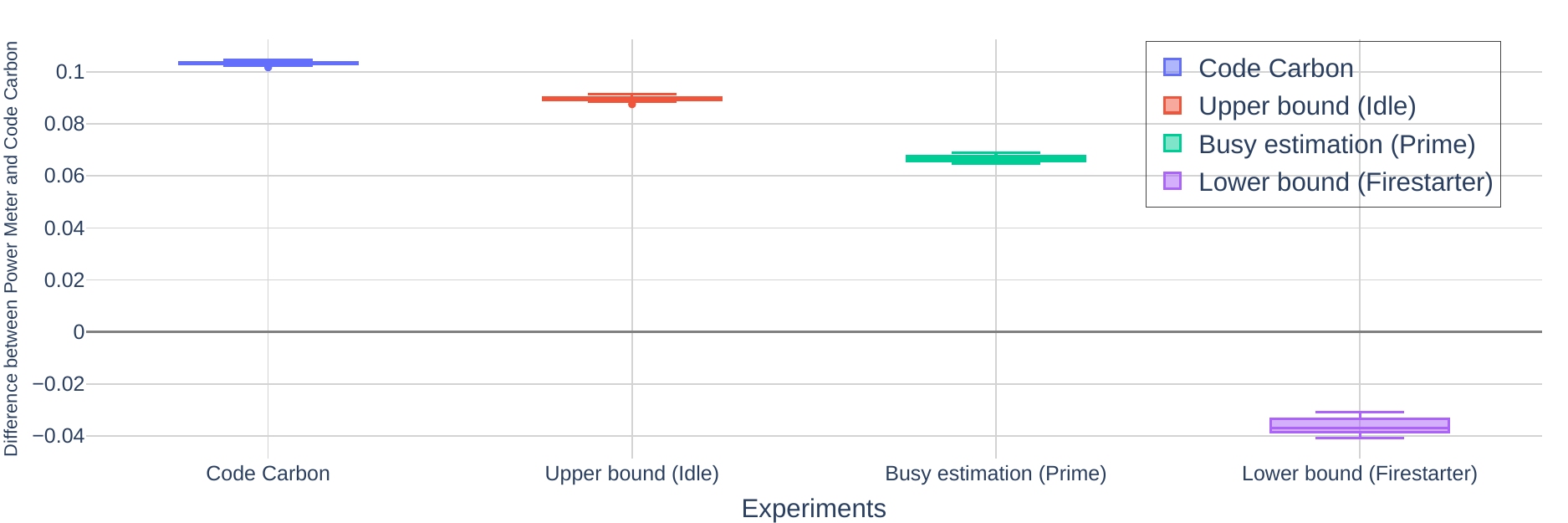}
    \caption{Boxplot with differences in energy measurements between Code Carbon and the power meter for single-GPU training.}
    \label{fig:box_code_carbon}
\end{figure*}
We highlight the results of our bounded approach compared to Code Carbon in \autoref{fig:box_code_carbon}. The base Code Carbon approach is $10.3\%$ off the power meter, while the bounds given are $\qty{8.9}{}$ to $\qty{-3.6}{\%}$. 
The load estimation via prime number calculation already improves the upper bound to $6.6\%$. Measurements also do not deviate much from the median.
The narrow range of error implies a constant off-socket load for EA-HAS-Bench, which could be estimated through small preliminary experiments. 
If a non-constant load is encountered, the upper and lower bound still guarantee a valid estimation. Precisely defining what load gives the best bounds for this correction will be explored within future work.

\section{Conclusions, limitations and future work}

In this work, we proposed 
design principles for energy-aware neural architecture search benchmarks and empirically evaluated the recently published EA-HAS-Bench based on them. Such benchmarks should be built upon
reliable power measurements, allow for a wide range of device usage, and report the holistic energy cost for the model. 
Our study encompasses multiple power measurement tools, including Nvidia \ac{SMI}, \ac{NVML} and Code Carbon, which we compared to an external power meter during the data collection phase of EA-HAS-Bench.

For this comparison, we 
determined the GPU power consumption from the
power meter with the help of calibration experiments and Intel's RAPL interface. We also identified a range of off-socket power consumptions caused by different loads on the CPU and memory.
The measurements made by Nvidia \ac{SMI} showed poor correlation to the power meter; 
we believe that this is caused by the sampling procedure using \ac{SMI} itself, producing sparsely sampled epochs. 
The faulty behaviour
was validated by a complementary study using the underlying \ac{NVML} library, 
in which the power measurements thus obtained were observed to show high correlation with those from the power meter.

We observed poor usage of the GPU during single-GPU training for the RegNet search space sampled in EA-HAS-Bench, with only up to $\qty{40}{\%}$ of the GPU being used at any given time. 
For single-GPU training, power consumption was found to correlate linearly with GPU usage and GPU memory usage, while for multi-GPU training, power consumption appeared to be only correlated to GPU memory usage.

We additionally proposed a method to improve the Code Carbon holistic energy reporting by compensating for the overestimation of memory consumption. To this end, we reuse the bounds we calculated for off-socket power consumption.
This provides a solid foundation for the reported energy costs based on empirically measured values rather than biased assumptions about memory consumption. 
In practice, these corrections do not require an external power meter, if the server supports power reporting through the baseboard management controller. Interested parties should contact their local system administrator and advocate for better reporting standards.

There are several limiting factors to our work. We only evaluate on one type of GPU attached to one specific server. This inherently introduces some bias towards high-end GPUs, such as the NVDIA H100 we tested.
We sample a relatively low number of architectures during our main
experiments. While the number of epochs sampled stays statistically relevant, 
we cannot guarantee the statistical significance for full training cycles. 
We argue that the large-scale validation experiment showing high correlation between precisely measured full-training energy consumption and the readings from the power meter justifies keeping the budget low for the main experiments.

In the future, energy-aware NAS benchmarks should sample from a more device-agnostic search space 
and provide transferability towards hardware-constrained devices. 
Furthermore, finding the off-socket load on a device a priori for tighter bounds during energy cost reporting is an interesting direction for further investigation.
We believe that energy-aware neural architecture search benchmarks should be based on trusted
data. With this study, we hope to have contributed to the establishment of best practices
for measuring energy-aware benchmarks.  More importantly, we enhance the reproducibility of the analysed EA-HAS-Bench and base holistic energy cost reporting tools, such as Code Carbon, on evidence-based foundations rather than assumptions. We are hopeful this will lead to better energy-aware benchmarks in the future.

\section*{Acknowledgment}
We thank Anja Jankovic for her feedback throughout the writing process, and Hadar Shavit for his help with designing an efficient experimental setup.
This work was supported by the German Federal Ministry of the Environment, Nature Conservation, Nuclear Safety and Consumer Protection (GreenAutoML4FAS project no. 67KI32007A).
Further, this work was supported by the Hessian Ministry of Higher Education, Research, Science and the Arts (HMWK; projects “The Third Wave of AI”) and the National High-Performance Computing project for Computational Engineering Sciences (NHR4CES). 
This work was performed as part of the Helmholtz 
School for Data Science in Life, Earth and Energy (HDS-LEE) and received funding from the Helmholtz Association of German Research Centres.
Holger Hoos gratefully acknowledges support through an Alexander-von-Humboldt Professorship in Artificial Intelligence.

\begin{acronym}
    \acro{PUE}{power usage effectiveness}
    \acro{PSU}{power supply unit}
    \acro{MSR}{model-specific register}
    \acro{RAPL}{Running Average Power Limit}
    \acro{SMI}{System Management Interface}
    \acro{NVML}{NVIDIA Management Library}
    \acro{eCDF}{empirical cumulative distribution function}
    \acro{NAS}{Neural Architecture Search}
\end{acronym}

\bibliographystyle{IEEEtran}
\bibliography{IEEEabrv,IEEE-conference-template-062824.bib}
\end{document}